\pdfoutput=1

\documentclass[11pt]{article}

\usepackage{acl}

\usepackage{times}
\usepackage{latexsym}
\usepackage{ragged2e}
\usepackage{tabularx}
\usepackage{placeins}

\usepackage[T1]{fontenc}

\usepackage[utf8]{inputenc}

\usepackage{microtype}

\usepackage{graphicx}

\usepackage{inconsolata}

\title{A Dataset for the Detection of Dehumanizing Language}

\author{Paul Engelmann \\
  IT University of Copenhagen \\
  \texttt{paen@itu.dk} \\\And
  Peter Brunsgaard Trolle \\
  IT University of Copenhagen \\
  \texttt{ptro@itu.dk} \\\And
  Christian Hardmeier \\
  IT University of Copenhagen \\
  \texttt{chrha@itu.dk} \\
}

\begin{document}
\maketitle
\begin{abstract}
Dehumanization is a mental process that enables the exclusion and ill treatment of a group of people. In this paper, we present two data sets of dehumanizing text, a large, automatically collected corpus and a smaller, manually annotated data set. Both data sets include a combination of political discourse and dialogue from movie subtitles. Our methods give us a broad and varied amount of dehumanization data to work with, enabling further exploratory analysis and automatic classification of dehumanization patterns. Both data sets will be publicly released.
\end{abstract}

\section{Introduction}
Dehumanization, the act of depicting someone as less than human, can be seen in many different examples, such as against African Americans \citep{mekawi2016white}, Arabs \citep{prati2016predicting} as well as between Israelis and Palestinians \citep{bruneau2017enemy}. Dehumanization can range from blatant to subtle forms of varying degrees \citep{bain2009attributing}, making automated, general detection difficult. \citet{mendelsohn2020framework} present one of the first computational works on dehumanization through explicit feature engineering, using lexicon and word embedding based approaches to detect dehumanizing associations across several years in a New York Times corpus. 
Outside of this, there is little computational work on dehumanization. We believe that the lack of work can be attributed to a vague general definition of dehumanization and a pronounced focus on content moderation, rather than the underlying processes of hateful content.

Additionally, we notice a lack of data sets specializing on dehumanization. While similar data, such as social media hate speech data \citep{silva2016analyzing, zhong2016content, DBLP:journals/corr/abs-2006-08328}, exists, these do not capture the specifics of dehumanization. 
Hate speech and dehumanization differ in the sense that hate speech is a surface phenomenon, representing the observable aspects of hateful content, whereas dehumanization describes the underlying attitude for certain types of hate speech.

As a result, we wish to provide two dehumanization focused data sets to allow work on general identification and detection of dehumanization. Both data sets are in English and collected from the OpenSubtitles \cite{lison2016opensubtitles2016} as well as the Common Crawl\footnote{Common Crawl: https://commoncrawl.org/} corpora. One data set consists of a larger, unlabelled corpus, while the other is an evaluation set consisting of human annotated samples, labelled by two independent annotators. Both data sets were extracted using keywords, which include target groups from ethnic, religious and sexual backgrounds, as well as common animal metaphor keywords and moral disgust terms from the Moral Foundations Dictionary\footnote{https://moralfoundations.org/other-materials/} \cite{graham2009liberals}.

For dehumanization patterns, we limit ourselves to patterns inspired by \citet{mendelsohn2020framework} and \citet{haslam2006}, where a sample is considered dehumanizing if it contains at least one of the following categories: negative evaluation of a target group, denial of agency, moral disgust, animal metaphors, objectification. Animal metaphors and objectification specifically relate to a human being compared to an animal or object with the intent to cause harm. 
\textit{Trigger Warning: This paper contains examples of hateful content that some may find upsetting.}
\section{Related Work}
Since computational work on dehumanization is sparse, we focus on related dehumanization research and other annotation efforts in fields such as hate speech detection. \citet{kteily2022dehumanization} provide an overview of current trends and challenges regarding dehumanization. \citet{mendelsohn2020framework} focus on the use of the NRC-VAD Lexicon \citep{mohammad2018obtaining}, which features 20,000 English keywords, rated by annotators based on their associated valence, dominance and arousal in the range of 0 to 1. Valence, in particular, describes the evaluation of an event or concept and assigns it a value, ranging from unpleasant to pleasant \citep{osgood1957measurement, russell1980circumplex}. \citeauthor{mendelsohn2020framework} hypothesise that low valence is an indication of potential dehumanization in the form of a negative evaluation of a target group, while low dominance suggests dehumanization in the form of denial of agency. These, together with word embeddings made out of combining several keywords for moral disgust and vermin metaphors, are leveraged to identify dehumanized target groups. \\
Examining hate speech data sets, \citet{mathew2021hatexplain} focus on explainable hate speech detection, aiming to increase the interpretability of hate speech detection models. \citet{DBLP:journals/corr/abs-1909-04251} provide a benchmark that not only tries to identify hate speech, but also expects generative models to be able to intervene in hateful discussions using automatically generated responses. \\
For automated abuse detection, \citet{DBLP:journals/corr/abs-1908-06024} provides an overview for several techniques and methods that are commonly employed. Transformer based models have shown particular promise in hate speech detection. An example is HateBERT \citep{caselli2020hatebert}, a BERT model trained from the ground up on hate speech data, outperforming the standard BERT model on the detection of hate speech. 

\section{Data Set Collection}
\subsection{OpenSubtitles}
OpenSubtitles \cite{lison2016opensubtitles2016} is a data set consisting of movie and TV series subtitles. It contains fictitious, high quality dialogue, curated by professional writers and thus possessing potentially more subtle dehumanization compared to standard dialogue. \\
We extract sentence windows with a size of 5 grammatical sentences per window, split based on quotation marks, under the condition that they contain at least one keyword from the religious, ethnic, sexual, moral disgust or animal category. A complete list of all keywords can be found in Table \ref{table:keywords}. To ensure that we do not over-represent a category, we limit each to 20\% of the samples. Since the data set can include multiple different subtitles for the same movie, deduplication and preprocessing has been performed, including replacement of URLs, identifiable names through a placeholder token, as well as transforming emojis into their equivalent text so that models may use them for inference. 
\subsection{Common Crawl}
The Common Crawl is an open repository of web crawled data, which features crawls from all over the internet. This data set allows us to take standard dialogue from everyday users, which allows us to extract more common dehumanization patterns. As the Common Crawl includes several petabytes of data in total, we have selectively extracted data from political forums, as political discourse is prone to the use of dehumanization \cite{cassese2021partisan}. As we limit ourselves to English, the Common Crawl data features discourse primarily focused on American and British politics. \\
Random web pages from these forums were extracted and preprocessed using jusText \cite{pomikalek2011justext}, to remove boilerplate code from the website. Additional preprocessing was performed similarly to the OpenSubtitles data. Examples for both can be found in Table \ref{table:corpus_examples}.
\begin{figure}
    \centering
    \includegraphics[width=\linewidth]{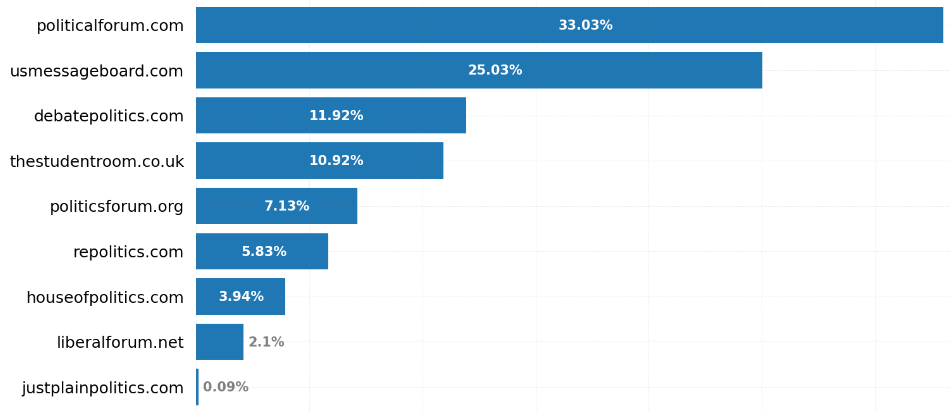}
    \caption{Percentage of tokens extracted from each forum in Common Crawl}
    \label{fig:cc_split}
\end{figure}
\subsection{Labelled Data Set}
The evaluation data set is a subset of the previously extracted data sources, thus containing the same limitations and processing steps as before. We extract 50\% of the data from OpenSubtitles and 50\% from Common Crawl, ensuring that each keyword group is equally likely from both sources. \\
The examples were labelled by two annotators. Each annotator was informed of the chosen criteria with their definition and artificial example sentences, which were not present in the data set. The example sentences can be found in Table \ref{table:annotator_examples}. An annotator could pick between the labels \textit{Yes, No, Not Sure} to signify if dehumanization is present. The category \textit{Not Sure} was reserved for cases where an annotator was not able to confidently pick an option, either due to missing context or ambiguous meaning of words. \\

\section{Analysis of the Data}

\begin{table}
    \centering
    \scalebox{1.0}{
        \begin{tabularx}{\linewidth}{ r |X }
        \hline
        Common Crawl & \RaggedRight Trump continually harps on violence from \textbf{[\textsc{ethnic group}]} in South American gangs, claims that \textbf{[\textsc{religious group}]} terrorists are in the caravan, that \textbf{[\textsc{ethnic group}]} are going to bring in diseases.\\ 
        \hline
        & \RaggedRight For many of us, this is revolting. Men dancing with men. \textbf{[\textsc{sexual group}]} in this country today break the law. \\
        \hline
        \hline
        OpenSubtitles &  \RaggedRight [...] Fuck it! I can't reason with a hairy, \textbf{[\textsc{ethnic group}]} [\textsc{slur}]. \\
        \hline
        &  \RaggedRight They do it in the back, in the butt. That's \textbf{gross}.\\ 
        \hline
        \end{tabularx}
    }
    \caption{Data examples, keyword matches are bolded}
    \label{table:corpus_examples}
\end{table}
\subsection{Unlabelled Data Set}
\label{sec:unlabelled}
A total of 565,304 paragraphs were extracted from both data sources, with 318,179 paragraphs being extracted from OpenSubtitles and 247,125 paragraphs from Common Crawl. We achieve a roughly equal split when considering tokens per corpus.
The Common Crawl part was created with data from the January 2021 crawl up to the October 2023 crawl. The contribution from each chosen forum can be found in Figure \ref{fig:cc_split}. \\
We tested two binary classifiers for the automatic detection of dehumanizing utterances. One is a baseline model, calculating the mean valence over each paragraph and using the previously chosen keywords as our criteria for dehumanization. The other is a fine tuned version of HateBERT, trained using the whole network, a learning rate of $5\cdot 10^{-5}$ and 4 epochs with 90\% of samples from the labelled set. A comparison between the baseline, HateBERTs from \citep{caselli2020hatebert} and our fine tuned version of HateBERT can be found in Table~\ref{table:BERTmetrics}. 
\begin{table}[ht]
    \centering
    \scalebox{0.85}{
        \begin{tabular}[width=\linewidth]{|l|cc|c|} 
        \hline
         & Precision & Recall & F1  \\
        \hline
        Baseline & 0.2402 & 0.7536 & 0.3643\\ 
        \hline
        HateBERT fine tuned & \textbf{0.6514} & 0.5462 & \textbf{0.5941}\\ 
        \hline
        HateBERT abuseval & 0.4825 & 0.5308 & 0.5055 \\ 
        \hline
        HateBERT hateval & 0.5833 & 0.1750 & 0.2692  \\
        \hline
        HateBERT offenseval & 0.3474 & \textbf{0.7615} & 0.4771 \\
        \hline
        \end{tabular}
    }
    \caption{Model metrics, evaluated on the labelled data set}
    \label{table:BERTmetrics}
\end{table}
The baseline identifies 10.6\% of data as dehumanizing, while HateBERT finds 8.03\% of data to be dehumanizing. Qualitative analysis of each approach with randomly selected samples show that the baseline identifies cases where dehumanization can be tied directly to the use of specific words, such as: \\
\textit{She left her} [\textsc{insult}] \textit{son here. Do you know my mother? His mother is a} [\textsc{sexual slur}]. \\
HateBERT finds more nuanced examples in the corpus: \\
\textit{[...] Just tell her you're not that into her anymore. [...] Ending a relationship is kind of like pulling off a bloodsucking leech.}\\
and in general detects negative animal metaphors, moral disgust as well as extremely negative evaluation of groups relating to ethnicity and sexuality. 

Using word2vec embeddings, with the same approach as \citep{mendelsohn2020framework}, we examine similarities between sexual keywords and moral disgust keywords. Results are compared to similarities with the label \textit{american}, as it is not limited to specific topics in our corpus, though we do not expect \textit{american} to be a neutral label due to the political bias in our data. \\
We achieve a significantly higher similarity with moral disgust for \textit{gay(s), lesbian, queer, transsexual(s), homosexual(s)} than \textit{american} (Wilcoxon's signed-rank test, $p < 0.05$). Examples include:  \\
\textit{I wouldn't even share a washing machine or a drinking fountain with that totally disgusting and disease ridden \textsc{[insult]} \textsc{[sexual group]}.} \\
Significance can not be established between sexual and animal keywords ($p > 0.05$). \\
For ethnic groups and animal keywords with the same comparison label, we achieve higher similarity with animals for \textit{african, russians, indian(s), mexican, korean, chinese} ($p < 0.05$). Examples include: \\
\textit{Cruelty is cruelty, whether the victim be a chicken or a malnourished \textsc{[ethnic group]}.}\\
No significant similarity for ethnic groups and moral disgust can be established however~($p~>~0.05$).

\subsection{Labelled Data Set}
\label{sec:labelled}
The labelled data set consists of 918 annotated samples, 450 of which were taken from Common Crawl and 468 from OpenSubtitles. These were excluded from the unlabelled data set. The labelling was performed independently and discussed after 600 samples. The other 318 samples were labelled without further discussion. For inter-annotator agreement using Krippendorff's alpha \citep{krippendorff2011computing}, we achieve a score of 0.4846 for samples before the discussion and a score of 0.4920 for samples after the discussion. Removing those cases where at least one annotator could not confidently answer \textit{Yes} or \textit{No}, we have a score of 0.5398 before the discussion and 0.5508 after the discussion. Related hate speech datasets \cite{sachdeva2022measuring} achieve a similar scoring, ranging from 0.5 to 0.6 for Krippendorff's alpha. From 55 positive annotations, that both annotators agree on, 41.8\% are animal metaphors, 29.09\% negative target evaluation, 10.90\% denial of agency and 9.09\% moral disgust and objectification.  
Examples of dehumanization for each pattern can be found in Table \ref{table:data_examples}. 

\begin{table}
    \centering
    \scalebox{0.9}{
        \begin{tabularx}{\linewidth}{ r |X }
        \hline
        Negat. Eval. of Group & \RaggedRight [\textsc{religious group}] don't whine? [\textsc{religious group}] INVENTED whining. [...]\\ 
        \hline
        Denial of Agency &  \RaggedRight [...] Keep remin[d]ing us how vacuous people become when they are as brainwashed as the Salem witch trial hooligans \\
        \hline
        Moral Disgust &  \RaggedRight I left the Dem party myself in 1998 after just six years in disgust [...]\\
        \hline
        Animal Metaphors & \RaggedRight [...] They very likely killed you, ya [\textsc{slur}] lab rat. \\ 
        \hline
        Objectification &  \RaggedRight He is poison. A pimple on a hogs ass.\\ 
        \hline
        \end{tabularx}
    }
    \caption{Examples from the labelled data set}
    \label{table:data_examples}
\end{table}

\section{Discussion}
As seen in Table \ref{table:BERTmetrics}, our HateBERT F1 score is quite low compared to other binary hate speech classification efforts. \citep{DBLP:journals/corr/abs-2006-08328} achieve a F1 score of 0.7713 using BERT. We believe that this is due to the low amount of data used for fine tuning and the fact that the patterns are not equally distributed, as seen in Section \ref{sec:labelled}, causing some of them to be under-represented. However, we believe that the analysis still gives a decent estimate of what can be expected from the data and that particularly common patterns of dehumanization, such as the use of animal metaphors, are frequently employed in both data sets. \\
For the labelled data set, we had to make several assumptions during the labelling process. 
Several of our samples include conversations about the event of someone being dehumanized. We did not recognize this as dehumanization, as we do not see the retelling of an event as possessing the same illocutionary force as direct dehumanization. Thus we restricted our labelling to those samples that included the author either being the target of dehumanization or dehumanizing someone else. \\
In about 0.5\% of our samples authors dehumanize themselves, for example through animal metaphors. We chose to label these as \textit{Not Sure}, as these do not directly target anyone with the intent to cause harm, but rather talk about hypothetical scenarios of dehumanization. In cases like these it was difficult to argue for or against dehumanization, since the intent to cause harm is not immediately clear. \\
Furthermore, the labelling process revealed several cases that highlight the requirement for specific domain knowledge to be able to accurately assess if someone is being dehumanized. Take the following example: \\
\textit{Newslime is the major reason Californians are making a mass exodus from the woke state. [...] Newslime is a white Obammy.}\\
Without knowing about the then governor of California, Gavin Newsom, it would be difficult to understand that he is being compared to slime, as \textit{Newslime} could also refer to someones real name. These cases showcase that it can be very difficult to detect dehumanization without having any kind of domain or context knowledge at hand and hints towards the direction that models may have to go to be able to perform effective detection.
\section{Conclusions}
Due to the ever evolving nature of dehumanization and abuse in general, automated detection methods stand before a significant challenge. We hope that by curating a dehumanization focused data set, we provide enough incentive for others to start exploring potential ways of developing computational dehumanization methods and tackle the fight against online abuse.

\section*{Limitations}
There exists an inherent bias in both data sets, as political discourse features a large amount of our data. We recognize that this might not be typical of other types of discourse. In particular, since we deal with political themes, dehumanization will focus on political topics and might not be able to translate well into general dehumanization detection. Since the data is in English and a lot of nuance is based on English grammar, we do not guarantee that the models trained on this data are generally able to detect dehumanizing speech in other languages. \\
Furthermore, keyword based extraction of large corpora always runs the risk of not being able to cover all potentially relevant keywords and thus missing out on data relevant for the task. This case is no different. We hope that we cover a wide enough spectrum of keywords, however these could always be expanded or further divided into subgroups to better differentiate between their attributes. 

\section*{Acknowledgements}
Peter Brunsgaard Trolle was supported by the European Union under grant agreement 101084457 (SafeNet). Views and opinions expressed are those of the authors and do not necessarily reflect those of the European Union. The European Union cannot be held responsible for them. \\
The authors acknowledge the IT University of Copenhagen HPC resources made available for conducting the research reported in this paper.

\bibliography{custom}

\begin{thebibliography}{23}
\expandafter\ifx\csname natexlab\endcsname\relax\def\natexlab#1{#1}\fi

\bibitem[{Bain et~al.(2009)Bain, Park, Kwok, and Haslam}]{bain2009attributing}
Paul Bain, Joonha Park, Christopher Kwok, and Nick Haslam. 2009.
\newblock Attributing human uniqueness and human nature to cultural groups: Distinct forms of subtle dehumanization.
\newblock \emph{Group Processes \& Intergroup Relations}, 12(6):789--805.

\bibitem[{Bruneau and Kteily(2017)}]{bruneau2017enemy}
Emile Bruneau and Nour Kteily. 2017.
\newblock The enemy as animal: Symmetric dehumanization during asymmetric warfare.
\newblock \emph{PloS one}, 12(7):e0181422.

\bibitem[{Caselli et~al.(2020)Caselli, Basile, Mitrovi{\'c}, and Granitzer}]{caselli2020hatebert}
Tommaso Caselli, Valerio Basile, Jelena Mitrovi{\'c}, and Michael Granitzer. 2020.
\newblock Hatebert: Retraining bert for abusive language detection in english.
\newblock \emph{arXiv preprint arXiv:2010.12472}.

\bibitem[{Cassese(2021)}]{cassese2021partisan}
Erin~C Cassese. 2021.
\newblock Partisan dehumanization in american politics.
\newblock \emph{Political Behavior}, 43:29--50.

\bibitem[{Graham et~al.(2009)Graham, Haidt, and Nosek}]{graham2009liberals}
Jesse Graham, Jonathan Haidt, and Brian~A Nosek. 2009.
\newblock Liberals and conservatives rely on different sets of moral foundations.
\newblock \emph{Journal of personality and social psychology}, 96(5):1029.

\bibitem[{Haslam(2006)}]{haslam2006}
Nick Haslam. 2006.
\newblock \href {https://doi.org/10.1207/s15327957pspr1003\_4} {Dehumanization: An integrative review}.
\newblock \emph{Personality and Social Psychology Review}, 10(3):252--264.
\newblock PMID: 16859440.

\bibitem[{Krippendorff(2011)}]{krippendorff2011computing}
Klaus Krippendorff. 2011.
\newblock Computing krippendorff's alpha-reliability.

\bibitem[{Kteily and Landry(2022)}]{kteily2022dehumanization}
Nour~S Kteily and Alexander~P Landry. 2022.
\newblock Dehumanization: Trends, insights, and challenges.
\newblock \emph{Trends in cognitive sciences}.

\bibitem[{Lison and Tiedemann(2016)}]{lison2016opensubtitles2016}
Pierre Lison and J{\"o}rg Tiedemann. 2016.
\newblock Opensubtitles2016: Extracting large parallel corpora from movie and tv subtitles.

\bibitem[{Mathew et~al.(2021)Mathew, Saha, Yimam, Biemann, Goyal, and Mukherjee}]{mathew2021hatexplain}
Binny Mathew, Punyajoy Saha, Seid~Muhie Yimam, Chris Biemann, Pawan Goyal, and Animesh Mukherjee. 2021.
\newblock Hatexplain: A benchmark dataset for explainable hate speech detection.
\newblock In \emph{Proceedings of the AAAI conference on artificial intelligence}, volume~35, pages 14867--14875.

\bibitem[{Mekawi et~al.(2016)Mekawi, Bresin, and Hunter}]{mekawi2016white}
Yara Mekawi, Konrad Bresin, and Carla~D Hunter. 2016.
\newblock White fear, dehumanization, and low empathy: Lethal combinations for shooting biases.
\newblock \emph{Cultural diversity and ethnic minority psychology}, 22(3):322.

\bibitem[{Mendelsohn et~al.(2020)Mendelsohn, Tsvetkov, and Jurafsky}]{mendelsohn2020framework}
Julia Mendelsohn, Yulia Tsvetkov, and Dan Jurafsky. 2020.
\newblock A framework for the computational linguistic analysis of dehumanization.
\newblock \emph{Frontiers in artificial intelligence}, 3:55.

\bibitem[{Mishra et~al.(2019)Mishra, Yannakoudakis, and Shutova}]{DBLP:journals/corr/abs-1908-06024}
Pushkar Mishra, Helen Yannakoudakis, and Ekaterina Shutova. 2019.
\newblock \href {http://arxiv.org/abs/1908.06024} {Tackling online abuse: {A} survey of automated abuse detection methods}.
\newblock \emph{CoRR}, abs/1908.06024.

\bibitem[{Mohammad(2018)}]{mohammad2018obtaining}
Saif Mohammad. 2018.
\newblock Obtaining reliable human ratings of valence, arousal, and dominance for 20,000 english words.
\newblock In \emph{Proceedings of the 56th annual meeting of the association for computational linguistics (volume 1: Long papers)}, pages 174--184.

\bibitem[{Mollas et~al.(2020)Mollas, Chrysopoulou, Karlos, and Tsoumakas}]{DBLP:journals/corr/abs-2006-08328}
Ioannis Mollas, Zoe Chrysopoulou, Stamatis Karlos, and Grigorios Tsoumakas. 2020.
\newblock \href {http://arxiv.org/abs/2006.08328} {{ETHOS:} an online hate speech detection dataset}.
\newblock \emph{CoRR}, abs/2006.08328.

\bibitem[{Osgood et~al.(1957)Osgood, Suci, and Tannenbaum}]{osgood1957measurement}
Charles~Egerton Osgood, George~J Suci, and Percy~H Tannenbaum. 1957.
\newblock \emph{The measurement of meaning}.
\newblock 47. University of Illinois press.

\bibitem[{Pomik{\'a}lek(2011)}]{pomikalek2011justext}
Jan Pomik{\'a}lek. 2011.
\newblock Justext.

\bibitem[{Prati et~al.(2016)Prati, Moscatelli, Pratto, and Rubini}]{prati2016predicting}
Francesca Prati, Silvia Moscatelli, Felicia Pratto, and Monica Rubini. 2016.
\newblock Predicting support for arabs' autonomy from social dominance: The role of identity complexity and dehumanization.
\newblock \emph{Political Psychology}, 37(2):293--301.

\bibitem[{Qian et~al.(2019)Qian, Bethke, Liu, Belding, and Wang}]{DBLP:journals/corr/abs-1909-04251}
Jing Qian, Anna Bethke, Yinyin Liu, Elizabeth~M. Belding, and William~Yang Wang. 2019.
\newblock \href {http://arxiv.org/abs/1909.04251} {A benchmark dataset for learning to intervene in online hate speech}.
\newblock \emph{CoRR}, abs/1909.04251.

\bibitem[{Russell(1980)}]{russell1980circumplex}
James~A Russell. 1980.
\newblock A circumplex model of affect.
\newblock \emph{Journal of personality and social psychology}, 39(6):1161.

\bibitem[{Sachdeva et~al.(2022)Sachdeva, Barreto, Bacon, Sahn, Von~Vacano, and Kennedy}]{sachdeva2022measuring}
Pratik Sachdeva, Renata Barreto, Geoff Bacon, Alexander Sahn, Claudia Von~Vacano, and Chris Kennedy. 2022.
\newblock The measuring hate speech corpus: Leveraging rasch measurement theory for data perspectivism.
\newblock In \emph{Proceedings of the 1st Workshop on Perspectivist Approaches to NLP@ LREC2022}, pages 83--94.

\bibitem[{Silva et~al.(2016)Silva, Mondal, Correa, Benevenuto, and Weber}]{silva2016analyzing}
Leandro Silva, Mainack Mondal, Denzil Correa, Fabr{\'\i}cio Benevenuto, and Ingmar Weber. 2016.
\newblock Analyzing the targets of hate in online social media.
\newblock In \emph{Proceedings of the International AAAI Conference on Web and Social Media}, volume~10, pages 687--690.

\bibitem[{Zhong et~al.(2016)Zhong, Li, Squicciarini, Rajtmajer, Griffin, Miller, and Caragea}]{zhong2016content}
Haoti Zhong, Hao Li, Anna~Cinzia Squicciarini, Sarah~Michele Rajtmajer, Christopher Griffin, David~J Miller, and Cornelia Caragea. 2016.
\newblock Content-driven detection of cyberbullying on the instagram social network.
\newblock In \emph{IJCAI}, volume~16, pages 3952--3958.

\end{thebibliography}

\appendix

\newpage
\section{Appendix}
\label{sec:appendix}

\begin{table}[ht]
    \centering
    \scalebox{0.75}{
        \begin{tabularx}{\linewidth}{ r |X }
        \hline
        Negat. Eval. of Group & \RaggedRight I really hate [\textsc{religious group}], nothing's worse than being close to one.\\ 
        \hline
        Denial of Agency &  \RaggedRight You are so stupid, you can't even think for yourself! \\ 
        \hline
        & \RaggedRight He is as dumb as a rock. \\ 
        \hline
        Moral Disgust &  \RaggedRight Men holding hands is gross. \\ 
        \hline
        & \RaggedRight Those two should never hang out together, she will be just as filthy as her! \\ 
        \hline
        Animal Metaphors & \RaggedRight Why do you look like a monkey? \\ 
        \hline
        & \RaggedRight All men are stupid sheep. \\ 
        \hline
        Objectification &  \RaggedRight He's nothing more than dirt. \\ 
        \hline
        & \RaggedRight All she does is [\textsc{sexual slur}] herself out.
        \end{tabularx}
    }
    \caption{Example Sentences for Annotators}
    \label{table:annotator_examples}
\end{table}

\begin{table}
    \centering
    \scalebox{0.7}{
        \begin{tabularx}{1.65\linewidth}{ r|r|r|r|r }
        \hline
        Religious & Ethnic & Sexual & Moral Disgust & Animals \\ 
        \hline
        muslim(s) & foreigner(s) & gay(s) & sin(s) & vermin \\
        jews(s) & immigrant(s) & lesbian(s) & sinned & parasite(s) \\
        christian(s) & white(s) & homosexual(s) & sinning & rodent(s) \\
        & black(s) & bisexual(s) & whore & rat(s) \\
        & american(s) & transgender(s) & impiety & mice \\
         & asian(s) & queer(s) & impious & cockroach(es) \\
         & indian(s) & lgbtq & gross & termite(s) \\
         & russian(s) & lgbtqia & tramp & bedbug(s) \\
         & african(s) & glbt & unchaste & fleas \\
         & arab(s) & lgbtqqia & intemperate & primate(s) \\
         & turkish & genderqueer & wanton & monkey(s) \\
         & hispanic(s) & genderfluid & profligate & ape(s) \\
         & latino(s) & intersex & trashy & gorilla(s) \\
         & mexican(s) & pansexual & lax & donkey(s) \\
         & chinese & transgender(s) & blemish & dog(s) \\
         & japanese & transsexual(s) & pervert(s) & snake(s) \\
         & korean(s) & transexual(s) & stain(s) & cow(s) \\
         &  & transvestite(s) & disgust* & lamb(s) \\
         &  & transgendered & deprav* & goat(s) \\
         &  & asexual & disease* & pig(s) \\
         &  & agender & unclean* & sheep*\\
         &  & aromantic & contagio* & chimp*\\
         &  & & indecen* & chick*\\
         &  & & sinful* & \\
         &  & & sinner* & \\
         &  & & slut* & \\
         &  & & dirt* & \\
         &  & & profan* & \\
         &  & & repuls* & \\
         &  & & sick* & \\
         &  & & promiscu* & \\
         &  & & lewd* & \\
         &  & & adulter* & \\
         &  & & debauche* & \\
         &  & & defile* & \\
         &  & & prostitut* & \\
         &  & & filth* & \\
         &  & & obscen* & \\
         &  & & taint* & \\
         &  & & tarnish* & \\
         &  & & debase* & \\
         &  & & desecrat* & \\
         &  & & wicked* & \\
         &  & & exploitat* & \\
         &  & & wretched* & \\
         
        \end{tabularx}
    }
    \caption{Complete keyword list, *-marked keywords are prefixes}
    \label{table:keywords}
\end{table}


\end{document}